%% file: paper.tex
\documentclass[journal]{IEEEtran}

\usepackage{amsmath,amsfonts,amssymb,amsthm,mathtools}
\usepackage{color}
\usepackage{stmaryrd}
\usepackage{multirow}
\usepackage[numbers]{natbib}
\usepackage{fix-cm}
\usepackage{url}
\usepackage{times}
\usepackage{graphicx}
\usepackage[export]{adjustbox}
\usepackage{latexsym}
\usepackage[linesnumbered,ruled,vlined]{algorithm2e}
\usepackage{algpseudocode}
\usepackage{caption}
\usepackage{subcaption}
\usepackage{soul}

\newcommand{\ti}{I}                         
\newcommand{\sti}{\bar{\imath}}             
\newcommand{\inst}{\mathcal I}              
\newcommand{\domn}{\mathcal D}              
\newcommand{\grnd}{\mathcal G}              

\theoremstyle{definition}
\newtheorem{defin}{Definition}

\definecolor{dkgreen}{rgb}{0,0.6,0}
\definecolor{gray}{rgb}{0.5,0.5,0.5}
\definecolor{mauve}{rgb}{0.58,0,0.82}
\definecolor{darkgray}{rgb}{0.2,0.2,0.2}
\definecolor{lightgray}{rgb}{0.7,0.7,0.7}

\newcommand{\stm}[3]{\left(\operatorname{#1},\operatorname{#2},\operatorname{#3}\right)}
\newcommand{\norm}[2]{\operatorname{#1}\left(#2\right)}
\newcommand{\name}[1]{$\operatorname{#1}$}
\newcommand{\namem}[1]{\operatorname{#1}}

\DeclareMathOperator*{\card}{card}
\DeclareMathOperator*{\dom}{dom}
\DeclareMathOperator*{\vals}{vals}

\makeatletter
\newcommand*{\rom}[1]{\expandafter\@slowromancap\romannumeral #1@}
\makeatother

\begin{document}

\title{Norms, Institutions, and Robots  
}



\author{%
  Stevan Tomic, %
  Federico Pecora
  and Alessandro Saffiotti%
\thanks{All authors are with the Center for Applied Autonomous Sensor
  Systems, School of Science and Technology, \"{O}rebro University,
  70187 \"{O}rebro, Sweden.  Contact author's e-mail:
  stevan.tomic@aass.oru.se.}%
}



\markboth{Journal of \LaTeX\ Class Files,~Vol.~14, No.~8, August~2015}%
{Shell \MakeLowercase{\textit{et al.}}: Bare Demo of IEEEtran.cls for IEEE Journals}

\maketitle

{\em \small This work has been submitted to the IEEE for possible publication. Copyright may be transferred without notice, after which this version may no longer be accessible.}

\begin{abstract}
  Interactions within human societies are usually regulated by social
  norms. 
  If robots are to be accepted into human society, it is essential that
  they are aware of and capable of reasoning about social norms.
  In this paper, we focus on how to represent social norms in
  societies with humans and robots, and how artificial agents such as
  robots can reason about social norms in order to plan appropriate
  behavior. We use the notion of institution as a way to formally
  define and encapsulate norms,
  and we provide a formal framework for institutions.
  Our framework borrows ideas from the field of multi-agent systems to
  define abstract normative models, and ideas from the field of robotics
  to define physical executions as state-space trajectories.  By
  bridging the two in a common model, our framework allows us to use the
  same abstract institution across physical domains and agent types.  We
  then make our framework computational via a reduction to CSP and show
  experiments where this reduction is used for norm verification,
  planning, and plan execution in a domain including a mixture of humans
  and robots.


\end{abstract}

\begin{IEEEkeywords}
Norms, Institutions, Institutional robotics, Social robotics, Mixed
human-robot society, Cognitive robotics
\end{IEEEkeywords}

\input{introduction}
\input{relatedWork}
\input{institution}
\input{domain}
\input{grounding}
\input{admisadher}
\input{computation}
\input{reasonExampleTrade}
\input{robotExample}
\input{conclusion}




\bibliographystyle{IEEEtranN}
\bibliography{biblio}   

%
%

\end{document}

%% file: introduction.tex
\section{Introduction}
\label{sec:intro}

\begin{quote}
  \textit{Roby needs a new battery.  He enters a hardware store, goes to
    the clerk, Sally, and they agree on the type of the battery and its
    price.  Then Roby handles some cash to Sally, and she gives him the
    battery.  Roby takes the battery and rolls out of the store.}
\end{quote}

%
%

\noindent
Roby, of course, is a robot.  Like most of us, both Roby and Sally
follow social norms in their interactions.  Social norms are a
fundamental part of our society: they influence our behavior in shops,
at work, on the road, when dining with friends, or when playing soccer.
Norms create prescriptions for social behaviors, facilitate decision
making and regulate how humans communicate, act and cooperate.  By
constraining our activities, norms make behavior more predictable.
Failure to cope with social norms causes difficulties to account for the
intentions behind one's behavior and can easily be evaluated as odd,
amusing, dumb, uncanny, or simply unlawful.

For robots to become part of our society, they will need to be aware of
social norms, reason about these norms, and act according to the
expectations that they create.  
%
%
A number of formal and computational models have been proposed in the
field of Multi-Agent Systems (MAS) to do so
\citep{article:BoellaNorms,weiss2013multiagent}, and we review many of
them in the next section.
As we shall see, however, these models are not fully adequate to be used
in physically embodied robotic systems, as they often lack a clear
operational semantics~\cite{alechina2013computational} and are typically
disconnected from physical execution~\cite{GrossiDignum.faabs2004}.


In this paper, we propose a framework for normative reasoning that can
be used by robots that operate in human society.  Our framework
contains formal elements to represent explicitly: (1) abstract norms,
that we encapsulate in reusable structures called \emph{institutions}:
these describe social situations and include roles, actions, artifacts,
as well as a set of norms that bind them; (2) a physical \emph{domain}
where these norms are to be applied; and (3) the relation between the
abstract norms and the physical domain, that we call \emph{grounding}.
The vignette above shows an instance of an abstract `trading'
institution instantiated in a physical hardware store, where the roles
of buyer and seller are grounded to Roby and Sally, and the artifacts
are grounded to the cash and the battery.

A key point of our framework is that it combines insights from the
fields of MAS and robotics.  From MAS, we borrow the separation
between abstract normative models and concrete realizations; from
robotics, we borrow the notions of physical execution and state-space
trajectories.  Our framework thus includes: (A)~A \textit{formal model
  of institutions} which encodes the relation between abstract norms and
their instantiation into a concrete robotic domain; (B)~a full
\textit{computational model} which enables norm verification, planning,
and plan execution by robots; and (C) support for ~\textit{artifacts},
which is of high importance in robotics since robots and humans interact
and coordinate via relevant objects in the environment.

Our framework defines institutions as abstractions that can be applied
to different sets of heterogeneous agents, including robots and humans,
which are not specifically built to work together.
%
%
%
%
This feature is especially important in the case of robots, 
%
%
%
since it would not be sensible, or economical, for a manufacturer to
design a robot to work within a particular institution
or to cater to specific norms, as this would restrict the social
contexts in which it can be used.
%
Robots should be able to reason about the roles they play in a
particular institution, the obligations that they have to fulfill, while
using corresponding artifacts to do so.  This spawns a need to create a
correspondence between a robot's domain (the concrete environment in
which it operates) and an institution (the current social context of the
robot).

This paper is organized as follows. Section~\ref{sec.realted} surveys
related work.  We introduce institutions and domains in
Sections~\ref{sec.inst} and~\ref{sec.dom}, respectively.  In
Section~\ref{sec.grounding} we relate domains and institutions, and show
how abstract norms can be given domain-specific semantics.  In
Section~\ref{sec.properties} we discuss the reasoning problems
associated with institutions, and in Section~\ref{sec.computation} we
define a computational model to address these problems.
Section~\ref{sec.reasoning_example} illustrates the model on a simple
trading example, and Section~\ref{sec.rexample} demonstrates it with
real robots and humans.  


%% file: relatedWork.tex
\section{Related Work}
\label{sec.realted}

%
Our analysis of the state of the art begins with an overview of the
general concept of norms.  An extensive overview of social norms in
the literature is given by \citet{article:BoellaNorms}, who observe
that research on norms is rooted in different areas, including
philosophy~\citep{alchourr1971normative} and
sociology~\citep{gibbs1965norms}.  It is noted that concepts and
theories from other disciplines should be used for normative
specifications in Computer Science.
%
%
In Computer Science, prominent work on norms was done
by~\citet{meyer1994deontic}, which led to deontic logic becoming a
dominant tool for modeling norms. The reason lies in the lack of other
methods to define behaviors which are illegal but nevertheless
possible, since illegal behavior is usually ruled out by problem
specification.
\citet{alechina2013computational} stress the importance of operational
semantics for normative languages, which is essential for creating
computational frameworks. The authors also discuss the lack of a
`one-size-fits-all' normative formalism.
%
%
%
%
%
As we will see, many of the specific choices made in our framework
stem from such general questions.
%


%
In our work, we use \emph{institutions} as a way to encapsulate norms and their context. In MAS
institutions are commonly described as a \textit{set of norms}~\citep{weiss2013multiagent}, or as ``rules of the game in the society''~\citep{north1990institutions}.
Frameworks supporting this idea include:
MOISE~\citep{hannoun2000moise}, 
OPERA~\citep{dignum2004model}, ISLANDER~\citep{Esteva20021045},
AMELI~\citep{esteva2004ameli}, InstAL~\citep{cliffe2006answer},
and JaCaMo\citep{boissier2013multi}.
They all abstract from the pure agent representation to a social level,
where the notion of \textit{roles} is usually defined. Roles are
associated with \textit{norms} and agents can enact roles, and act
depending on the specified norms.  The Agents and Artifacts (A\&A)
framework~\citep{ricci2007give} also models environmental
\textit{artifacts} to address objects in a working environment, as it is
important for physical interactions. The social level is typically
associated with the level on which agents operate (in the further text
referred to as a \textit{domain}).  This is usually done via the
\textit{count-as} principle, which is related to our notion of
grounding.
%
%
It should be noted that the problem of operationalization of organizational rules and connecting them to concrete execution in a domain is not trivial task to achieve. Thus, most of frameworks concerned with abstraction cannot verify/enforce formal properties. Others do not offer full abstraction from a concrete (physical) domain in a level of details required for robot operations. Most also lack means to equip agents with planning and plan execution mechanisms.  As we shall see, our framework addresses these issues.



A line of research focuses on the questions of how norms can be used
operationally via reasoning.
%
%
%
The groundwork addressing the operationalization of norms is done by
\citet{oren2009towards}.
They are able to track changing state of norm by extending
its deontic representation. Authors introduce conditions for norms
activation, violation, maintenance, etc.
Frameworks that define operational semantics are typically based on
these ideas.
%
For example, ~\citet{alvarez2011normative} use their framework for
monitoring norm compliance. The norm operational semantics is realized
by translating norm conditions to rules and then, by using a production
system, they compile rules to structures that are used to describe the
state of the norms.
A similar approach is done by ~\citet{garcia2009constraint}, where their
production system is enhanced with constraint-satisfaction techniques,
which leads to more expressiveness of norms, as claimed by the
authors. Also, as they claim, one of the limitations of their language
is the inability to plan.
%
%
%
In general various other frameworks use operational semantics for
monitoring norms compliance. 
For instance, \citet{bolton2013generating} present experiments where the behavior of humans is observed and its adherence to norms is verified. 
However, since they do not explicitly
address the planning problem and it is not clear how they can be
extended to autonomous planning, they are omitted in this summary of
related work.


Alongside norm monitoring, the operational semantics of
norms is also used to execute agents actions.
%
%
An approach for specifying and executing normative protocols comes from
~\citet{artikis2009specifying}. They make norms computational by
producing transition systems via action language (\textit{C+}), which
can be used to execute agent actions.  However, as shown, the approach
is not practical for run-time agent execution due to its long
compilation time for action descriptions.
%
%
Work closely related to this paper is done
by~\citet{alvarez2016thesis}.
%
%
While being focused on Service-Oriented Architectures, their framework
is general with respect to the type of agent.  The authors extend
deontic logic to dyadic deontic logic (to support conditions) and to LTL
(to support temporal reasoning).  They reduce the operational semantics
of norms to fluent-based semantics and further translate it into PDDL,
therefore bringing norm monitoring close to a planning problem.
%
%
LTL supports only qualitative temporal relations, while in our
work, we are additionally interested in quantifiable temporal
relations so we can provide a sufficient level of specification and
reasoning in human-robot interactions.

%
For cyber-physical (social) systems in general or human-robot
interaction in particular, it is important that norms semantics can be
used for real-time reasoning about physical aspects of the environment.
This requires, but is not limited to, qualitative and quantitative
temporal and spatial reasoning for monitoring norm compliance and
planing robotic action to adhere to norms.  Furthermore, norms semantics
should contain extended count-as principle to include objects in the
environment and map them to a social level (as artifacts).
Those are some of the issues we address in our approach.
%
%
%
%
Several frameworks deal with sanctioning mechanisms.  We do not address
this aspect in our framework since its importance for robotics is not
clear, but leave it for future work.

Some works in human-robot interaction, do not use the notion of norms explicitly, rather, ontologies that capture whole aspects of a situation. \citet{quintas2018toward} focus on overcoming uncertainty by probabilistic plans, however not providing a general formalism that can verify if execution adheres to the provided model. \citet{wang2015formalization} deal with verification and the generic aspect of behavior, however, focus on behavior couplings and do not deal with context and dependence of behaviors on objects in the environment. All of these aspects are addressed in our work. 

Regarding related work about norms for robots, \citet{can2016making}
post strong arguments about why we should provide social norms to
robots. They argue that robots should behave in human-like fashion and
this is only possible if robots can organize and coordinate their
behaviors with the social expectations of others.
%
%
Norms in automated planning have been studied
by~\citet{panagiotidi2013reasoning}, who extended the STRIPS language
to include norm semantics. A goal state is then a state of the world
where the effects of all active norms are achieved.
\citet{CirilloEtAl.tist10}, \citet{montreuil2007planning},
\citet{pecora2012constraint} extend planning to support human activities.
\citet{kockemann2014grandpa} have developed the notion of
\textit{interaction constraints} relating to robot actions and human
activities, thus allowing for norm-aware plans to be generated. In our
previous work~\citep{TooCoolForSchool}, we have introduced Socially
Aware Planner (SAP) that supports \textit{social
  norms}. With social norms we were able to relate the current social
context to concrete robots and humans activities.
%
%
In general, while several authors have accounted for normative behavior
in robots, none of them provide a means to represent norms in an
abstract and reusable way that does not depend on a domain.
A step in this direction was made by \citet{Carlucci2015}, 
and the work in this paper provides one further step.


Institutions have received much less attention in the field of Robotics
than norms.  An institutional framework for robots was developed by
\citet{Pereira2014}.
Institutions are defined in terms of Petri Nets, which gives the
framework a sound mathematical foundation but falls short of
including task planning.
``Institutional Robotics'' is an important work by \citet{Silva2015} in
which they discuss and analyze to great extend the meaning, importance,
and different ways of usage of institutions for robotics.  Their
experiments are based on \citet{Pereira2014}.

%% file: institution.tex
\section{Norms and Institutions}
\label{sec.inst}

%

The ingredients that define our institutions at an abstract level are a
set of \emph{artifacts}, a set of \emph{roles}, a set of \emph{actions},
and a set of {\em norms} that link roles, actions, and artifacts.
If we take as an example the game of football (soccer), artifacts
include a ball, a field and two goals; roles include goalkeeper, player,
referee and audience; actions include defending, scoring, and attacking;
and norms regulate how actions are performed, e.g., ``a \emph{player}
should \emph{attack} the opposite \emph{goal}'', or ``a
\emph{goalkeeper} can \emph{handle} the \emph{ball} while in the
\emph{penalty area}''.

More formally, let these sets be:
\begin{align*}
  \textit{Arts} &= \{art_1, art_2, \dots , art_a\} \\ 
  \textit{Roles} &= \{role_1, role_2, \dots , role_m\} \\ 
  \textit{Acts} &= \{act_1, act_2, \dots , act_k\}
\end{align*}


\noindent We define a \emph{normative statement}, or simply \emph{norm},
to be a predication over ground statements, where a ground statement is
a triple including a subject, a predicate, and an object.

\begin{defin} A {\em norm} has the form $q(stm^*)$, where
  $q$ is a qualifier and $stm^*$ are triples of the form:
  \begin{align*}
    stm \in \textit{Roles} \times \textit{Acts} \times (\textit{Arts} \cup \textit{Roles})
  \end{align*}
\label{norm.def}
\end{defin}


%
%
Qualifiers are deontic verbs, like \emph{must} or \emph{must-not}, or
relations, like \emph{inside} or \emph{before}.
By talking about ground statements, qualifiers define the normative
language of an institution.
For example, a unary qualifier for necessity can be used to express an
obligation like ``a goal-keeper {\em must} defend its goal''.
Qualifiers defined over pairs of statements can, for example, express
temporal concepts such as `before', `during', etc.  In
Section~\ref{sec.semantics} we shall see how these intuitive semantics
are formalized.  We distinguish between \emph{obligation norms}, that
impose obligatory actions, and \emph{modal norms}, that describe other
requirements on statements.


An obligation norm has a unary qualifier denoting that the action in the
statement must be executed.  The qualifier could indicate that it is
necessary to execute the action at least once, or that it should be
executed repeatedly.  An obligation norm may express that for example
that ``a goal-keeper {\em must} defend its goal'' as
$\norm{must}{\stm{goal-keeper}{defend}{ownGoal}}$, or that ``a buyer
{\em must} pay a seller'' as $\norm{must}{\stm{buyer}{pay}{seller}}$.
One can also define norms that state which actions are forbidden in an
institution using a unary qualifier \emph{must not}, e.g.,
$\norm{mustNot}{\stm{referee}{play}{footballField}}$.

A modal norm can encode where and when actions should be carried out, for
instance, the fact that ``a player plays {\em inside} a football-field''
can be represented as
$\norm{inside}{\stm{player}{plays}{footballField}}$.
A modal norms can have an $n$-ary qualifier, which can be used to specify
a relation between two or more statements, like in
$\norm{before}{\stm{buyer}{pays}{money},\stm{seller}{gives}{goods}}$,
representing ``the buyer pays {\em before} the seller gives the goods''.


Obligation norms and modal norms predicate over statements.  We define a
third type of norm that predicates over roles, namely, indicating the
normative minimum and the maximum number of agents that can enact (play) a
certain role, e.g., ``there can be only one goalkeeper (per team)''.

\begin{defin} A {\em cardinality norm} associates roles to the minimum and
  maximum cardinality:
\begin{align*}
\card : \textit{Roles} \to \mathbb{N} \times \mathbb{N}
\end{align*}
\end{defin}

We can now define an institution as follows:

\begin{defin} An {\em institution} is a tuple
\begin{align*}
\inst = \langle \textit{Arts}, \textit{Roles}, \textit{Acts}, \textit{Norms} \rangle,
\end{align*}
where
$\textit{Norms} = \textit{OBN} \cup \textit{MON} \cup \{\card\}$
is a collection of obligation, modal and cardinality norms.
\end{defin}
%

The above formalization is in line with the literature in social and
economic sciences, where institutions are typically seen as mechanisms
that regulate social action by defining and upholding
norms~\citep{ostrom2009understanding}.  It is also in agreement with
\citet{north1990institutions}, who sees institutions as containers of
the ``rules of the game in a society'',
and with \citet{harre1972explanation}, who stress the importance of
roles, seen as ``normative concept[s], focusing on what is proper for a
person in a particular category to do''.

%% file: domain.tex
\section{Robots and Domains}
\label{sec.dom}

An institution is an abstraction, which can be instantiated in different
concrete systems that are physically different but have the same
organizational structure.  For instance, the same football institution
can be used to regulate a game played by a group of children and one
played by a group of robots.  We model such a concrete system through
the notion of a \emph{domain}.

\begin{defin}
\label{def:domain}
A {\em domain} is a tuple $\mathcal{D} = \langle A, O, B, F, R \rangle$, where
\begin{itemize}
\item $A$ is a set of {\em agents},
\item $O$ is a set of {\em physical entities},
\item $B$ is a set of {\em behaviors},
\item $F \subseteq A \times B \times (O \cup A)$ is a set of {\em affordances},
\item $R$ is a finite set of {\em state variables}.
\end{itemize}
%
\end{defin}

\noindent The agents $A$ could be a mix of humans and robots, e.g.,
\begin{align*}
\{\namem{Tom}, \namem{Sally}, \namem{Ann}, \namem{Nao3}, \namem{Roomba1}, \namem{Turtlebot4} \}.
\end{align*}
$B$ is the set of all behaviors that agents can execute, e.g.,
\begin{align*}
\{ \namem{walk}, \namem{play}, \namem{talk}, \namem{dance}, \namem{run} \}.
\end{align*}
Physical entities $O$ are ordinary objects in the domain, e.g.,
\begin{align*}
\{ \namem{whiteboard}, \namem{ball}, \namem{floor}, \namem{chair},
\namem{meadow}, \namem{brush} \}. 
\end{align*}
Agents can be heterogeneous and have different capabilities.  The
affordance relation $F$ indicates which agents can execute which
behaviors with which object, e.g.,
\begin{align*}
\{ \stm{Sally}{walk}{floor}, \stm{Nao3}{play}{ball},\\
 \stm{Roomba1}{clean}{brush} \}.
\end{align*}
The state variables $R$ define properties of the entities in
the domain. They may indicate the position of an object, the size of an
agent, the status of activation of a behavior, etc.
%
For instance,
$\rho = \namem{active}(\namem{walk},\namem{Nao3})$
is a state variable that indicates whether the \name{walk} behavior is
active on the agent \name{Nao3}. We denote with $\vals(\rho)$ the set of
possbile values of state variable $\rho$, e.g.,
$\vals(\namem{active}(\namem{walk},\namem{Nao3})) = \{\top, \bot \}$.
Some other state variable $\namem{pos}(\namem{ball1})$, for instance,
can indicate the qualitative position of an object:
%
%
\begin{align*}
\vals(\namem{pos}(\namem{ball1})) = \{\namem{back-half},
\namem{front-half}, \namem{goal-area}, \dots \}. 
\end{align*}
In a different domain, the \name{pos} state variable may hold continuous
values in a given coordinate system.

\begin{defin}
\label{def:stspace}
Given a domain $\domn = \langle A,O,B,F,R \rangle$, the \emph{state
  space} of $\domn$ is $\mathcal{S} = \prod_{\rho \in R} \vals(\rho)$.
The value of $\rho$ in state $s \in S$ is denoted $\rho(s)$.
\end{defin}

In a dynamic environment, 
the values of most properties change over time.
In our formalization, we represent time points by natural numbers in
$\mathbb{N}$, and time intervals by intervals $\ti = [t_1, t_2]$ of
$\mathbb{N}$.  We denote by $\mathbb{I}$ the set of all such time
intervals.
We then represent the evolution of properties over time by trajectories
of states.

%

\begin{defin}
\label{def:traje}
A \emph{trajectory} is a pair $(\ti,\tau)$, where $\ti \in \mathbb{I}$
is a time interval and $\tau : \ti \to \mathcal{S}$ maps time to states.
\end{defin}

\begin{figure}
\centering
\includegraphics[width=0.7\linewidth]{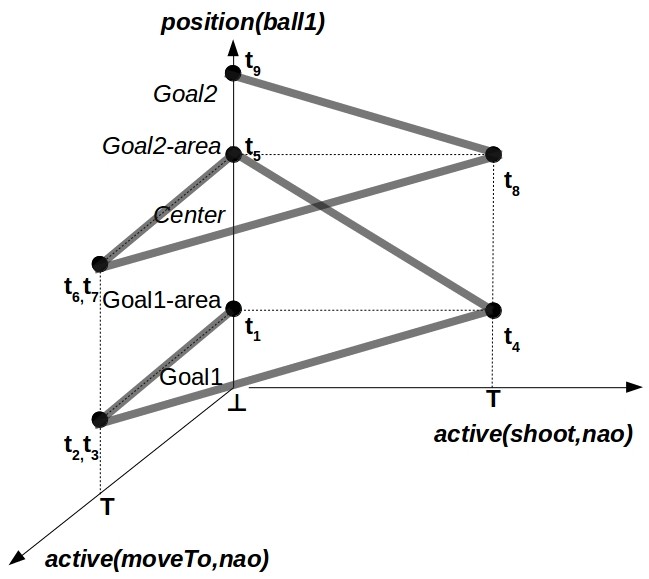}
\caption{An example of a trajectory visualized in the state space. It
  consists of 9 time points and describes an attacking sequence of
  agent \name{Nao}.}
\label{fig:traj}
\end{figure}

Figure~\ref{fig:traj} shows an example of a trajectory 
in the space defined by three state variables:
$\namem{position}(\namem{ball1})$,
$\namem{active}(\namem{moveTo},\namem{Nao})$ and
$\namem{active}(\namem{shoot},\namem{Nao})$.  The agent \name{Nao}
engages in behaviors \name{moveTo} and \name{shoot} until the ball is in
the goal.  At time $t_1$ the ball is at \name{goal1Area}.  At $t_2$ and
$t_3$ the ball is in the same position, and the agent's behavior
\name{moveTo} is active:
$\namem{active}(\namem{moveTo},\namem{Nao})(\tau(t_3)) = \top$.  At
$t_4$ \name{moveTo} finishes and \name{Nao} shoots the ball:
$\namem{active}(\namem{shoot},\namem{Nao})(\tau(t_4)) = \top$.  At $t_5$
the ball is in a new position:
$\namem{position}(\namem{ball1})(\tau(t_5)) = \namem{goal2Area}$.  A
similar sequence repeats until the ball's position is \name{Goal2}.





%% file: grounding.tex
\section{Relating Institutions and Domains}
\label{sec.grounding}

The above formalization separates all aspects of the institution
abstraction from the physical domain.  This separation provides the key
to reuse the same abstract institution to describe or regulate different
systems of robots and humans.  We now study how to bind an institution
to a given domain.

\subsection{Grounding Institutions}

Consider the above football institution, and imagine a domain consisting
of a group of children in a meadow.  If the children want to play
football, they need to map physical objects and agents to the entities in
the football institution.  In other words, they need to \emph{ground}
the institution.


\begin{defin}
  Given an institution $\inst$ and a domain $\domn$, a \emph{grounding}
  of $\inst$ into $\domn$ is a tuple
  $\grnd = \langle \grnd_A, \grnd_B, \grnd_O \rangle$, 
  where:
  \begin{itemize}
  \item $\grnd_A \subseteq Roles \times A$ is a \emph{role grounding},
  \item $\grnd_B \subseteq Acts \times B$ is an \emph{action grounding},
  \item $\grnd_O \subseteq Arts \times O$ is an \emph{artifact grounding}.
  \end{itemize}
\end{defin}
We denote by $A_{role} = \{a \mid a \in A \wedge (role,a) \in {\mathcal
  G}_A\}$ the set of all agents to which a specific $role$ is grounded.
We define $B_{act}$ and $O_{art}$ in a similar way.

\begin{figure}
\centering
\includegraphics[width=\linewidth]{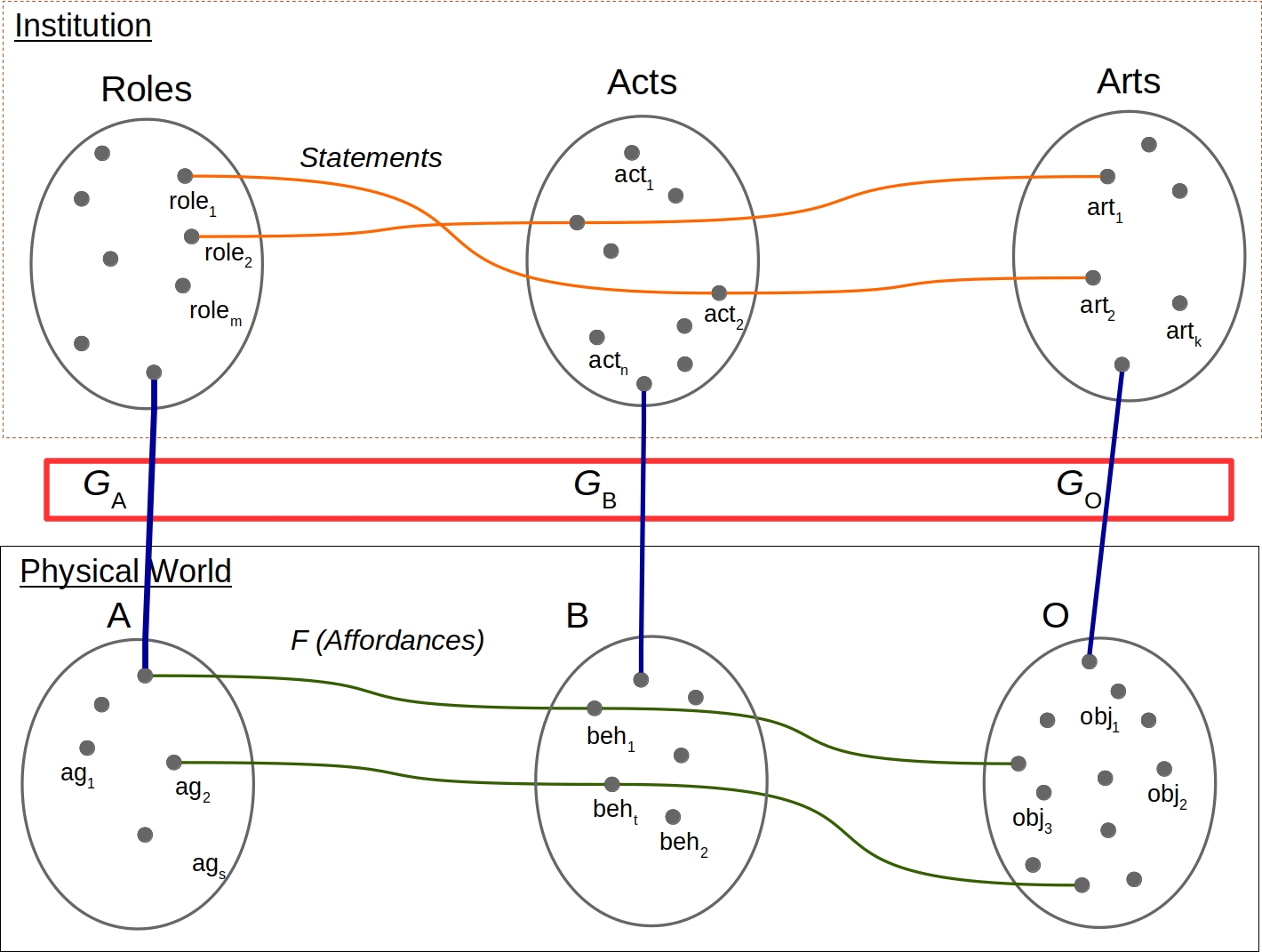}
\caption{The institution model, with a domain and grounding}
\label{fig:setInst}
\end{figure}

In our example, the children may decide to use two trees to ground the
two goal posts, and the meadow to ground the football field.  This
corresponds to artifact grounding, $\mathcal{G}_O$.  Institutional
artifacts, like the football field, are related to concrete objects in
the real world, like the meadow: $(\namem{meadow},\namem{footballField})
\in \mathcal{G}_O$.  The children also decide which role each child is
going to play.  This corresponds to role grounding, $\mathcal{G}_A$,
where, e.g., the role of goal-keeper is assigned to a child named Tom:
$(\namem{goalkeeper},\namem{Tom}) \in \mathcal{G}_A$.  If children can
dynamically join or leave the game, then $\mathcal{G}_A$ needs to be
changed dynamically.  Finally, specific behaviors of children, like
shooting or blocking the ball, can be used to realize institutional
actions, like attacking or defending. This is modeled by the action
grounding $\mathcal{G}_B$, e.g., $(\namem{Attack}, \namem{shoot}) \in
\mathcal{G}_B$.  Figure~\ref{fig:setInst} graphically represents the
relation between institution, domain, and grounding.


The grounding of an institution to a domain has some important
philosophical connotations.  The grounding $\mathcal{G}$ gives a certain
institutional meaning (a \textit{status}) to domain elements.  This fact
is known as the \emph{count-as} principle~\citep{jones1996formal}.
\citet{searle2005institution} explain this principle as assigning
`status functions' to elements, so that \textit{X counts as Y in
  context C}.  For example, a specific piece of paper counts as money in
a trading institution,
while putting a paper in the box could count as voting in an election
institution.  
Behaviors that are in the `count as' relationship with
institutional actions are also referred to as having institutional
\textit{power}~\citep{jones1996formal}. 



\subsection{Giving Semantics to Norm}
\label{sec.semantics}

Once we have decided a domain and a grounding, the syntactic elements
(roles, actions, and artifacts) of an institution acquire meaning in the
physical world.  However, this is not yet the case for norms: what is
the meaning of ``$\norm{must}{\stm{buyer}{pay}{goods}}$'' in a
particular domain?

We give semantics to norms in terms of trajectories in the state space.
For instance, the semantics of the above norm can be defined as all
trajectories where the action \name{pay} is executed at least once. The
semantics of the norm
$\norm{before}{\stm{buyer}{pays}{money},\stm{buyer}{takes}{goods}}$ can
be defined as all trajectories where the behavior that is grounded to
the \name{pay} action happens before the behavior that is grounded to
the \name{takes} action.
Formally, let $\mathbb{T}$ be the set of all possible trajectories
$(\ti,\tau)$ over the state-variables in domain $\mathcal{D}$. Let
$q(stm*)$ be any norm, as per Definition~\ref{norm.def}.  We define the
semantics of $q(stm*)$, written $\llbracket q(stm*) \rrbracket$, as:
%
%
\begin{align*}
  \llbracket q(stm*) \rrbracket \subseteq \mathbb{T}
\end{align*}
Intuitively, a norm identifies a set of trajectories, namely, those that
correspond to executions that comply with that norm.  In other words,
norms express constraints on the possible values that state variables
take over time.



In general, whenever we ground an institution $\inst$ to a domain
$\domn$, we must make sure that all of the qualifiers used to define
norms in $\inst$ are given semantics in terms of acceptable executions
in $\domn$.  To illustrate this concept, below, we provide examples of
semantics for a selection of qualifiers.  It is important to note that
these examples are not an exhaustive list of the qualifiers and norms
that can be encoded in our framework: they only serve to illustrate the
kinds of semantics that one can define over state variables in the
domain.

\subsection{Examples of Semantics}
\label{sec.norms}

In these examples, we assume that a domain $\domn = \langle A,O,B,F,R
\rangle$ and a grounding $\mathcal{G} = \langle \mathcal{G}_A,
\mathcal{G}_B, \mathcal{G}_O \rangle$ are given.
We also assume that the set of state variables $R$ in $\mathcal{D}$
contains state variable $\namem{active}(b,ag)$ for every pair of $(b,ag)
\in B \times A$, with values $\vals(active(b,ag)) = \{ \top, \bot \}$,
and agent $ag$ executes behavior $b$ in state $s$ iff
$\namem{active}(b,ag)(s) = \top$.
%


\textbf{Must.} The semantics of a `must' norm makes sure that 
certain behavior is active at least once in the given trajectory:
{\small
\begin{equation}
\label{norm.must}
\begin{split}
   \llbracket & \norm{must}{(role,act,art)} \rrbracket \equiv\\
   & \{ (\ti,\tau) \mid \forall a \in A_{role} . \exists (b,t) \in B_{act} \times \ti : \\
   & \namem{active}(b,a)(\tau(t)) = \top \}.
\end{split}
\end{equation}
}
%
%
The semantics of this norm is defined as all trajectories $({\mathbb
  I},\tau)$, where for each agent playing a $role$, there is at least
one behavior in the grounding of $act$ that is active for that agent at
some time $t$.  Variants of this semantics are possible, e.g., a
\emph{persistent} version, where the behavior should be enacted at all
times (not just once), or a version imposing that the artifact $art$
should be used.
%

\textbf{At.} This is an example of a norm with a spatial semantics:
{\small
\begin{equation}
\label{norm.at}
\begin{split}
  \llbracket & \norm{at}{(role,act,art)} \rrbracket \equiv \\
  \{ &(\ti,\tau) \mid \forall (b,a,t) \in B_{act} \times A_{role} \times \ti . \exists o \in O_{art}: \\
     &\namem{active}(b,a)(\tau(t)) = \top \implies \\
     &\namem{pos}(b,a)(\tau(t)) = \namem{pos}(o)(\tau(t)) \}.
\end{split}
\end{equation}
}
All agents in the role grounding for which behaviors grounded by the
action are active should have the same position as the position of at
least one artifact in the artifact grounding. This simple semantic
model can, of course, be changed to encode more sophisticated spatial
relations, e.g., the object has to be within certain boundaries of
another object.

\textbf{Use.} This norm regulates the usage of artifacts:
{\small
\begin{equation}
\label{norm.use}
\begin{split}
  \llbracket & \norm{use}{(role,act,art)} \rrbracket \equiv\\
  \{ &(\ti,\tau) \mid \forall (b,a,t) \in B_{act} \times A_{role} \times \ti . \exists o \in O_{art} : \\
     &\namem{active}(b,a)(\tau(t)) = \top \implies \namem{using}(b,a)(\tau(t)) = o \}.
\end{split}
\end{equation}
}
Similarly to the semantics of {\em at}, this states that all agents in
the role grounding with behaviors in the action grounding that are
active should use an object in the artifact grounding.




\textbf{Before.} This is an example of a norm that relates two
statements, and that has a temporal semantics:
%
%
{\small
\begin{equation}
\label{norm.before}
\begin{split}
  \llbracket & \norm{before}{(role_1,act_1,art_1),(role_2,act_2,art_2)} \rrbracket \equiv\\
          \{ &(\ti,\tau) \mid \forall
          \left(a_1 \in A_{role_1}, a_2 \in A_{role_2},
          b_1 \in B_{act_1}, b_2 \in B_{act_2}\right) \\
             &\namem{active}(b_1,a_1)(\tau(t_1)) = \top \wedge \namem{active}(b_2,a_2)(\tau(t_2)) = \top\\
             &\implies t_1 < t_2 \}
\end{split}
\end{equation}
}
This states that all behaviors in the action grounding, if active, have
to be in a certain order: the first should precede the second.  The
semantic model could be even more specific, addressing exact objects
with which actions are performed. In a similar way, it is also possible
to model the other qualitative temporal relations in Allen's Interval
Algebra~\citep{Allen.ai1984}.



%% file: admisadher.tex
\section{Reasoning about Institutions}
\label{sec.properties}

Together, grounding and semantics provide the connection between the
abstract elements in an institution and the physical entities in a
domain.  In our opening example of a trading institution, grounding is
what binds Roby to the ``buyer'' role and Sally to the ``seller'' role;
and the norm
$\norm{before}{\stm{buyer}{pays}{money},\stm{seller}{gives}{goods}}$,
induces, through its semantics, a temporal relation between the
behaviors executed by Roby and those executed by Sally.
Since the institution, the domain and their connection are all formal
elements in our framework, a robot can reason about them.
We now define two properties that are crucial in such reasoning.

%
%



\subsection{Admissibility}
\label{subsec.admissibility}

There are a few things we expect from an intuitively `good' grounding.
First, if an institution includes an obligation norm for some role, then
the agents to which that role is grounded should be capable of executing
the actions required by the norm.  More precisely, each such agent
should be capable of executing at least one behavior grounded to that
action using an object grounded to the relevant artifact.
Consider for example a football institution $\inst$ that includes a
(persistent) obligation norm
$\norm{must}{\stm{Goalkeeper}{Defend}{OwnGoal}}$.  Let $\grnd$ be a
grounding for $\inst$ such that $(\namem{Goalkeeper},\namem{nao}) \in
\grnd_A$, $(\namem{Defend},\namem{block}) \in \grnd_B$ and
$(\namem{OwnGoal},\namem{net1}) \in \grnd_O$.  We expect that
$\namem{nao}$ can use $\namem{block}$ behavior on $\namem{net1}$.
If this is not the case, then $\grnd$ should not be used as a grounding.
The following definition formalizes this intuition:
%
%
\begin{defin}
\label{def:executable1}
Let $\grnd$ be a grounding of $\inst$ into $\domn$.  An obligation norm
$q(role,act,art)$ in $\inst$ is \textit{executable} under $\grnd$ iff
\begin{align*}
  \forall a \in A_{role} . \exists (b,o) \in B_{act} \times O_{art} : (a,b,o) \in F.
\end{align*}
\end{defin}

A similar condition can be stated for obligation norms that refer to
actions whose object is another role: for example, an escort institution
might include an obligation norm like \textit{``the follower must follow
  the leader''}.
\begin{defin}
\label{def:executable2}
Let $\grnd$ be a grounding of $\inst$ into $\domn$.  An obligation norm
$q(role,act,role_o)$ in $\inst$ is \textit{executable} under $\grnd$ iff
\begin{align*}
  \forall a \in A_{role} . \exists (b,a_o) \in B_{act} \times A_{role_o}: (a,b,a_o) \in F.
\end{align*}
\end{defin}


Finally, a grounding should respect the cardinality of norms.
\begin{defin}
\label{def:card}
Let $\grnd$ be a grounding of $\inst$ into $\domn$.  A cardinality norm
$\card$ in $\inst$ is \emph{satisfied} for role $role \in \textit{Role}$
iff
\begin{align*}
  \min\left(\card(role)\right) \leq \left|A_{role}\right| \leq \max\left(\card(role)\right).
\end{align*}
\end{defin}

Putting these conditions together:
%
%
\begin{defin}
\label{def:admissible}
Let $\grnd$ be a grounding of $\inst$ into $\domn$.  $\grnd$ is
\emph{admissible} iff all its obligation norms are executable and its
cardinality norm is satisfied for all roles.
\end{defin}



\subsection{Adherence}
\label{subsec.adherence}

The admissibility property of grounding is not concerned with semantics,
nor with the dynamic aspects of the domain: with this property alone, we
cannot tell whether or not a given execution satisfies the norms in an
institution.  Consider the trading institution in our opening example:
the grounding of the roles and artifacts to our agents (Roby and Sally)
and to our objects (the cash and the battery) is admissible, which means
for instance that Sally has the capability to give a battery to Roby.
But admissibility alone does not discriminate an execution where Sally
does give the battery to Roby after she has got the cash, from an
execution where she does not.

In our framework, dynamic aspects are captured by trajectories.
Given an institution, admissible grounding and norm semantics, a
trajectory may or may not adhere to the institution's norms.  
The fact that a given trajectory satisfies these norms depends on the
specific semantics of the norms.
%
%
The following definition captures this intuition.

%

\begin{defin}
\label{def:adheres}
Let $\inst = \langle \textit{Arts}, \textit{Roles}, \textit{Acts},
\textit{Norms} \rangle$ be an institution and $\domn$ a domain.  Let
$\grnd$ be an admissible grounding of $\inst$ into $\domn$, and let
$\llbracket \cdot \rrbracket$ be a semantic function.  A trajectory
$(\ti,\tau)$ in $\domn$ \emph{adheres} to $\inst$ under $\grnd$ and
$\llbracket \cdot \rrbracket$ if
\begin{equation*}
  (\ti,\tau) \in \llbracket \namem{norm} \rrbracket, \forall
  \namem{norm} \in \textit{Norms}.
\end{equation*}
\end{defin}

In other words, a trajectory $(\ti,\tau)$ adheres to the institution
$\inst$ if it does not violate any of the constraints induced by the
institution's norms through the semantic function $\llbracket \cdot
\rrbracket$ during the time interval $\ti$ on which the trajectory is
defined.

\subsection{Reasoning Problems}
\label{sec.reasoningproblems}

An institution, grounding, trajectory, domain, and semantics could be
known (given); or some or all of these elements may have to be inferred
or calculated.  Different combinations of what is given in a particular
situation lead to different kinds of reasoning problems.
Table~\ref{table:problems} helps us identify and classify interesting
problems in a systematic way.

\paragraph{Verification} Ensure that the given trajectory
\textit{adheres} to the given institution with grounding, domain, and
semantics.
For example, a trajectory representing our initial trading vignette
would pass the verification test, whereas one where Sally runs away with
the money would not.



\paragraph{Grounding} Find an \textit{admissible} grounding for a given
$\inst$ and $\domn$.
For instance, a football player must be able to grab a ball in order to
be assigned to the role of a goalkeeper.


\paragraph{Planning} Generate a trajectory in domain $\domn$ that is
adherent to institution $\inst$, with a given grounding and semantics.
For Roby in our scenario, this is the problem of regulating its give and
take behaviors so that the norms of the trading institution are
satisfied.  Sometimes the Grounding and Planning problem may be
combined, that is, the planner has to choose agents, behaviors, and
objects, and to generate an adherent trajectory with them.


\paragraph{Recognition} The task here is to recognize which agents,
behaviors and objects are bound to which institutional elements.  For
example, what is the role of a person standing in front of a goal during
a football game?  Or what action is being executed by kicking a ball
into a net?

\paragraph{Relational Learning} We can identify the relational learning
(RL) problem when the task is to find normative relations or semantics
from a given trajectory in a given domain for a given institution and
grounding. That is, we need to answer questions like ``what does it
mean to \name{use} a \name{paymentUnit}?'', or more precisely, what
are the rules regulating how an agent (a buyer) should behave to
satisfy the norm $\norm{use}{\stm{buyer}{pay}{paymentUnit}}$?

\paragraph{Institution Learning/Recognition} Institution learning is
the problem of learning institution structure by observing the
dynamics of agent interactions. Institution recognition addresses the
case where institutions are already defined, and the problem is to
recognize which institution is taking place and how it is grounded
within the observed domain. For example, if two persons are playing
with a ball, are they playing a game of football, basketball or
tennis?



\begin{table}[!ht]
  \caption{Overview of reasoning problems with institutions. Legend:
    $\top =$ given, $x =$ to be found, $- =$ not of interest.} 
  \label{table:problems}
  \centering
  \begin{tabular}{||c|c|c|c|c|c||}
    \hline
    {\em Problem} & $\inst$ & $\mathcal{G}$ & $(\tau,\ti)$ & $\mathcal{D}$ & $\llbracket \cdot \rrbracket$ \\ \hline\hline
    Verification  & $\top$ & $\top$ & $\top$ & $ \top $ & $ \top $  \\ \hline
    Grounding     & $\top$ & x      & $-$    & $ \top $ & $ \top $  \\ \hline
    Planning      & $\top$ & $\top$ &  x     & $ \top $ & $ \top $  \\ \hline
    Recognition   & $\top$ &  x     & $\top$ & $ \top $ & $ \top $  \\ \hline
    Relational learning           & $\top$ & $\top$ & $\top$ & $ \top $ &  x        \\ \hline
    Institution learning/recognition  &  x     &  x     & $\top$ & $ \top $ & $ \top $  \\ \hline
  \end{tabular}
\end{table}

%
%
%
%
%
%


%% file: computation.tex
\section {Computation}
\label{sec.computation}


We address the question of how to effectively compute a solution to the
reasoning problems presented in Table~\ref{table:problems}.  We focus
our attention on the verification problem (row \rom{1}) since this is a
basic step needed in all other problems.  We show that the verification
problem can be reduced to a known decision problem by introducing a
specific assumption.  We then discuss how verification can be used to
enable planning.


\subsection{Formulation as Constraint Satisfaction Problem (CSP)}
\label{subsec.formulationCSP}


We show that an institution grounded in a specific domain can be
naturally expressed as a collection of constraints over suitably chosen
state variables.  Specifically, given an institution $\inst$, domain
$\domn$, grounding $\grnd$, and semantic function $\llbracket \cdot
\rrbracket$, we can construct a constraint network $(W,\mathcal{C})$
where $W$ is a set of variables and $\mathcal{C}$ is a set of
constraints over $W$ defined as follows.
For each state variable $\rho \in R$, we introduce a variable $w_{\rho}
\in W$ whose domain consists of all possible bounded trajectories for
$\rho$.  Formally,
$\dom(w_{\rho}) = \{ (\ti, \tau) \mid \ti \in \mathbb{I}, \tau : \ti \to
\vals(\rho) \}$.
As for the constraints $\mathcal{C}$, we note that there is a one-to-one
correspondence between these constraints and the norm semantics as
defined in Section~\ref{sec.semantics} since those semantics limit the
possible trajectories in the state space.  Norm semantics are defined
over state variables $R$, while the constraints $\mathcal{C}$ are
defined over the corresponding variables in $W$.  For instance, the
obligation norm $n = \norm{must}{(role,act,art)}$ with the semantics
given in (\ref{norm.must}) above induces the following constraint in
$\mathcal{C}$:
\begin{align*}
  C_n \equiv 
  \bigwedge_{a \in A_{role}}
  \bigvee_{b \in B_{act}} 
  \left(w_{\namem{active}(b,a)} \diamond \top \right).
\end{align*}
%
%
where we use the notation $w_{\rho} \diamond v$ to indicate that the
value of $w_{\rho}$ is a trajectory such that $\rho$ assumes the value
$v$ at some point of it.  Formally:
\[
 w_{\rho} \diamond v \equiv
 \left(w_{\rho} = (\ti, \tau) \wedge \exists t \in \ti : \tau(t) = v \right).
\]
An assignment of values to the variables $W$ that satisfies all
constraints in $\mathcal{C}$ thus represents a particular trajectory
that adheres to the institution. Thus, the problem of finding an
adherent trajectory in an institution is reduced to the Constraint
Satisfaction Problem (CSP)~\citep{books/daglib/0076790}.


A trajectory defines values for all state-variables. It is, therefore,
possible to represent a trajectory as a collection of unary
constraints over variables. Given trajectory $(\ti,\tau)$ in
the state space defined by $R$, we can obtain a set of unary
constraints
$\mathcal{C}_{(\ti,\tau)} = \cup_{\rho \in R} C_{\rho}$, 
where
\begin{align*}
C_{\rho} \equiv
 \left( w_{\rho}(t) = \rho(\tau(t)), \forall t \in \ti \right).
\end{align*}

The constraint network $(W,\mathcal{C})$ represents the institution,
grounding, and semantics, while $(W,{\mathcal C}_{(\ti,\tau)})$
represents a given trajectory $(\ti,\tau)$.  Thus, the constraint
network
\begin{align*}
  (W, \mathcal{C} \cup \mathcal{C}_{(\ti,\tau)})
\end{align*}
has a solution if and only if trajectory $(\ti,\tau)$ adheres to the
institution with given grounding and semantics.  This addresses the
\emph{verification problem} listed in Table~\ref{table:problems}.  

This computational model is clearly too complex to be practical, as
variables may take on values representing any trajectory in state space.
In the next section, we put several assumptions in place which make the
verification problem feasible. 



\subsection{Solving the Verification Problem}


Constraints in the above representation
have to be checked at each time point in the interval $\ti$ of a given
trajectory. To keep the computational problem feasible, it is
reasonable to make some assumptions on how state variable values can
evolve. Henceforth, we assume a \emph{piece-wise constant
  temporal function} for trajectories. Hence, constraint checking need
not consider each time point in $\ti$, rather each contiguous interval
for which state variables have constant values.
This assumption is commonly made in temporal planning as well as
scheduling, as it allows to reason about the temporal sub-problem via
temporal constraint reasoning methods like Simple Temporal
Problems~\citep{DechterMP91}.  Timeline-based planning approaches use
this assumption to reduce the planning problem to that of constructing
trajectories in state
space~\citep{frank2003constraint,cesta2008unifying}.  A similar
assumption is made for integrating planning and
scheduling~\citep{ghallab1994representation}, and hybrid-reasoning for
robots~\citep{mansouri2016robot}. In all these approaches, the variables
in the underlying CSP represent flexible intervals of time within which
a state variable assumes a constant value.


In a verification problem the trajectory is given, that is, the values
of state variables over time are known.  We can use this to construct a
trajectory-specific constraint network as follows.  Let $\inst$ be an
institution linked to a domain $\domn$ via grounding $\grnd$ and
semantics $\llbracket \cdot \rrbracket$, and let $(\ti,\tau)$ be a
trajectory for the state variables $R$ in $\domn$.  Assume that
$(\ti,\tau)$ is such that each state variable $\rho \in R$ takes on one
of a finite set of values in $\vals(\rho)$, that is, for all $t \in
\ti$, $\rho(\tau(t)) \in \{\bar{v}_1, \dots , \bar{v}_{d}\}$.
Let $\mathcal{I}_\rho = \{ \sti_1, \dots , \sti_{e} \}$ be the set of
maximal sub-intervals $\sti$ of $\ti$ such that $\rho$ is constant over
$\sti$, that is, $\forall t\in \sti, \rho(\tau(t)) = \bar{v}_k$ for some
$k$.
We construct a constraint network $(\overline{W}, \mathcal{C})$ where
the variables are all the constant segments in the given trajectory,
that is,
$\overline{W} = \{ w_{\rho,\sti} \mid \rho \in R, \sti \in
\mathcal{I}_\rho \}$.
%


%
The constraints $\mathcal{C}$ are still derived directly from norm
semantics, however, the scope of the constraints now includes all of the
sub-variables $w_{\rho,\bar{I}}$ of a given state variable. For
instance,
the semantics of the norms $n_1 = \norm{must}{\stm{role}{act}{art}}$,
$n_2 = \norm{at}{\stm{role}{act}{art}}$ and $n_3 =
\norm{use}{\stm{role}{act}{art}}$ may lead, respectively, to the
following constraints:
\begin{align}
\label{const.must}
  C_{1} \equiv &
  \bigwedge_{a \in A_{role}}
  \bigvee_{b \in B_{act}} 
  \exists \sti : w_{\namem{active}(b,a),\sti} = \top\,,
\\
\label{const.at}
  C_{2} \equiv &
  \bigwedge_{a \in A_{role}}
  \bigwedge_{b \in B_{act}} 
  \bigvee_{o \in O_{art}} 
  \forall \sti :  w_{\namem{active}(b,a),\sti} 
  \\
  & \rightarrow 
  \left( \exists \sti' : \sti \sqsubseteq \sti'
  \wedge
  w_{\namem{pos}(b,a),\sti} = w_{\namem{pos}(o),\sti'}\right)\,,
  \notag
\\
\label{const.use}
  C_{3} \equiv &
  \bigwedge_{a \in A_{role}}
  \bigwedge_{b \in B_{act}} 
  \bigvee_{o \in O_{art}}
  \forall \sti : w_{\namem{active}(b,a),\sti} 
  \\
  & \rightarrow 
  \left( \exists \sti' : \sti \sqsubseteq \sti'
  \wedge
  (w_{\namem{using}(b,a),\sti'} = o) \right)\,.
  \notag
\end{align}
where $\sqsubseteq$ denotes interval inclusion.

\subsection{Planning via Verification}
\label{subsec.planning}

Planning requires to compute a trajectory that adheres to an
institution's norms under given semantics.  The CSP reduction shown
above is appropriate for verifying candidate plans such as those
considered by timeline-based planning approaches, which search the space
of possible `timelines' of state variables.  The collection of these
timelines is typically represented exactly as we have done above, in the
form of a constraint network with as many variables as there are
constant-valued intervals of time.  These planners employ constraint
reasoning techniques to verify that a candidate set of timelines (i.e.,
a candidate plan) adheres to constraints given in a domain
specification.  Some approaches, such as the one we use in
Section~\ref{sec.rexample} below, provide very expressive domain
specification languages, which include temporal, spatial, resource and
other constructs~\citep{mansouri2016constraint}. This will allow us to
express the semantics of norms directly in the domain specification, and
to leverage the planner's ability to search in the space of possible
timelines to find an adherent trajectory $(\ti,\tau)$.

%% file: reasonExampleTrade.tex
\section{Reasoning Example}
\label{sec.reasoning_example}

In this section, we unfold the concepts described so far on a trading
institution inspired by the vignette of Roby and Sally seen in the
Introduction.
%
%
The institution consists of two roles, a buyer and a seller. The buyer
pays with some form of payment and receives the purchased goods.  The
seller receives the payment and gives the goods to the buyer. In our
framework, these concepts are specified as follows.
{\small
\begin{align*}
  \textit{Roles} &= \{ \namem{Buyer}, \namem{Seller} \} \\ 
  \textit{Acts} &= \{ \namem{Pay}, \namem{ReceiveGoods}, \namem{ReceivePayment}, \namem{GiveGoods} \} \\
  \textit{Arts} &= \{ \namem{PayForm}, \namem{Goods} \}\\
\end{align*}
}
%
A set of obligations norms enforce that buyers pay and sellers give
goods, respectively:
{\small
\begin{align*}
  N1 = & \norm{must}{\stm{Buyer}{Pay}{PayForm}},\\
  N2 = & \norm{must}{\stm{Buyer}{ReceiveGoods}{Goods}},\\
  N3 = & \norm{must}{\stm{Seller}{ReceivePayment}{PayForm}},\\
  N4 = & \norm{must}{\stm{Seller}{GiveGoods}{Goods}}.
\end{align*}
}
Other norms regulate the transaction in time and usage of artifacts:
{\small
\begin{align*}
  N5 = & \norm{use}{\stm{Buyer}{Pay}{PayForm}},\\
  N6 = & \norm{use}{\stm{Buyer}{ReceiveGoods}{Goods}},\\
  N7 = & \norm{use}{\stm{Seller}{ReceivePayment}{PayForm}},\\
  N8 = & \norm{use}{\stm{Seller}{GiveGoods}{Goods}} \\
  N9 = & \norm{before}{\stm{Buyer}{Pay}{PayForm},\right. \\
        & \hspace{8ex}\left.\stm{Buyer}{ReceiveGoods}{Goods}} \\
%
%
%
%
\end{align*}
}
A cardinality norm ensures that there is exactly one buyer and one
seller in any trading instance:
{\small
\begin{align*}
\card(\namem{Buyer}) = (1,1), \card(\namem{Seller}) = (1,1).
\end{align*}
}

Let us now consider a concrete realization by providing a specific
domain $\domn = \langle A,B,O,F,R \rangle$.
{\small
\begin{align*}
  A = \{&\namem{Roby}, \namem{Sally}, \dots \} \\
  B = \{&\namem{give}, \namem{take}, \dots \} \\
  O = \{&\namem{cash}, \namem{battery}, \dots \} \\
  F = \{&\stm{Roby}{give}{cash}, \stm{Roby}{give}{battery}, \\
  &\stm{Roby}{take}{cash}, \stm{Roby}{take}{battery}, \\
  &\stm{Sally}{give}{cash},\stm{Sally}{give}{battery}, \\
  &\stm{Sally}{take}{cash}, \stm{Sally}{take}{battery}, \dots \} \\
  R = \{&\namem{active}(\namem{take},\namem{Roby}), \namem{active}(\namem{give}, \namem{Roby}), \\
        &\namem{active}(\namem{take},\namem{Sally}), \namem{active}(\namem{give},\namem{Sally}), \\
        &\namem{using}(\namem{take},\namem{Roby}), \namem{using}(\namem{give},\namem{Roby}), \\
        &\namem{using}(\namem{take},\namem{Sally}), \namem{using}(\namem{give},\namem{Sally}) \}
\end{align*}
}
$\domn$ may also include additional elements that are not relevant to
our story, e.g., an object `motor' and the fact that Sally has the
affordance ``(Sally, give, motor)''.
The state variable \name{active} indicates whether a given agent is
executing a given behavior (value = $\top$) or not (value = $\bot$),
while the value of the variable \name{using} is the name of the object
being used.

Finally, let's consider the following grounding $\grnd$:
{\small
\begin{align*}
  \grnd_A = \{& (\namem{Buyer}, \namem{Roby}), (\namem{Seller}, \namem{Sally}) \} \\
  \grnd_B = \{& (\namem{Pay}, \namem{give}), (\namem{ReceiveGoods}, \namem{take}), \\
              & (\namem{ReceivePayment}, \namem{take}), (\namem{GiveGoods}, \namem{give}) \} \\
  \grnd_O = \{& (\namem{PayForm}, \namem{cash}), (\namem{Goods}, \namem{battery}) \}
\end{align*}
}

\subsection{Admissibility}
\label{subsec.reason_admiss}

Recall that if a grounding $\grnd$ is admissible it comply to the obligation norms.  That is, each agent
grounded to a role mentioned in an obligation norm should have an
affordance with behavior and object which are grounded to the action
and artifact of that norm \--- see Definition~\ref{def:executable1}.
This condition is checked by Algorithm~\ref{alg:executable}.

\begin{algorithm}
  \SetEndCharOfAlgoLine{}
  \SetKwFunction{true}{true}
  \SetKwFunction{false}{false}
  \SetKwInOut{Input}{Input}
  \SetKwInOut{Output}{Output}
  \Input{$\norm{must}{(role,act,art)} \in \inst, \mathcal{D}, \mathcal{G}$}
  \Output{\true iff $r$ is executable}
  \SetKwFunction{exec}{Executable}
  \SetKwFunction{cap}{Capable}
  \SetKwProg{proc}{Procedure}{}{}
  \proc{\exec{$\norm{r}{(role,act,art)},\mathcal{D},\mathcal{G}$}}{
    \For{$(role,ag) \in \mathcal{G}.\mathcal{G}_A$}{
      \If{$\neg$ \cap{$ag,act,art,\mathcal{D},\mathcal{G}$}}{\Return \false\;}
    }
    \Return \true\;
  }
  \proc{\cap{$ag,act,art,\mathcal{D},\mathcal{G}$}}{
    \For{$(act,b) \in \mathcal{G}.\mathcal{G}_B$}{
      \For{$(art,o) \in \mathcal{G}.\mathcal{G}_O$}{
        \If {$(ag,b,o) \in \mathcal{D}.F$}{
          \Return \true\;
        }
      }
    }
    \Return \false;
  }
\caption {Executable}
\label{alg:executable}
\end{algorithm}

\noindent Procedure {\tt Executable} checks if an obligation norm is
executable by checking if all grounded agents are capable of executing
the required $act$ with the corresponding $art$.  The {\tt Capable}
procedure ensures that the object part of the statements can be used by
required actions grounded to corresponding behaviors.  The procedure is
run for all obligation norms to ensure overall admissibility
(Definition~\ref{def:admissible}).  For norm $n_1$, since
$(\namem{Buyer},\namem{Roby}) \in \grnd_A$ (line 2), the procedure
checks if \name{Roby} is capable of executing action \name{Pay} with the
artifact \name{PayForm}.  Since $(\namem{Pay},\namem{give}) \in \grnd_B$
(line 7), $(\namem{PayForm},\namem{cash}) \in \grnd_O$ (line 8), the
procedure verifies that $(\namem{Roby},\namem{give},\namem{cash}) \in F$
(line 9).  This is the case, so norm $n_1$ is deemed executable.  All
other obligation norms are verified in such a manner, and all are found
executable given the domain and grounding in this example.  Also, the
number of grounded agents is in the limits of the cardinality norm (not
shown in the algorithm), thus the grounding $\grnd$ is admissible.
 
\subsection{Adherence}

\begin{figure}
\includegraphics[width=\linewidth]{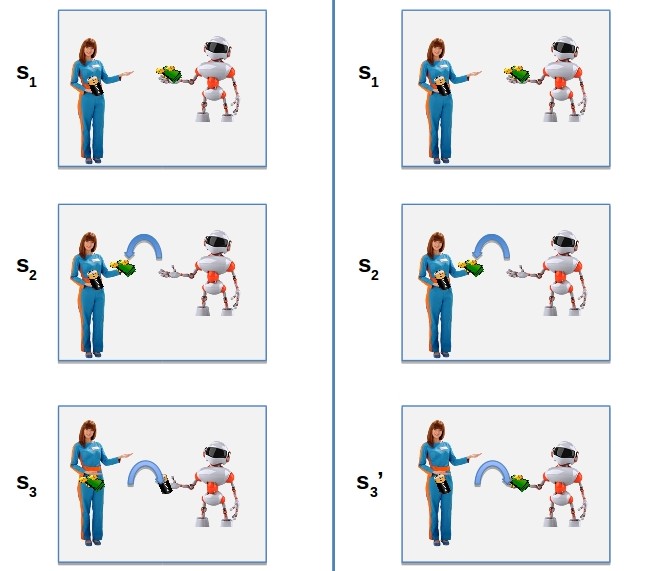}
\caption{Pictorial representation of two stories (trajectories).  The
  left one adheres to a trading institution, the right one doesn't.}
\label{fig:trade}
\end{figure}

Two similar trajectories are graphically summarized in
Figure~\ref{fig:trade}, where each picture shows a state.  
%
%
These states can be described as follows, where we omit the values of
inactive behaviors:
%
%
{\small
\begin{align*}
  s_1 = \{& {\namem{active}(\namem{give},\namem{Roby})}(t_2) = \top, \\
  &{\namem{using}(\namem{give},\namem{Roby})}(t_2) = \namem{cash} \} \\
  s_2 = \{& {\namem{active}(\namem{give},\namem{Roby})}(t_3) = \top, \\
  &{\namem{using}(\namem{give},\namem{Roby})}(t_3) = \namem{cash}, \\
  &{\namem{active}(\namem{take}, \namem{Sally})}(t_3) = \top, \\
  &{\namem{using}(\namem{take}, \namem{Sally})}(t_3) = \namem{cash} \} \\
  s_3 = \{& {\namem{active}(\namem{take},\namem{Roby})}(t_4) = \top, \\
  &{\namem{using}(\namem{take},\namem{Roby})}(t_4) = \namem{battery}, \\
  &{\namem{active}(\namem{give},\namem{Sally})}(t_4) = \top, \\
  &{\namem{using}(\namem{give}, \namem{Sally})}(t_4) = \namem{battery} \}  \\
  s_3' = \{& {\namem{active}(\namem{take},\namem{Roby})}(t_4) = \top, \\
  &{\namem{using}(\namem{take},\namem{Roby})}(t_4) = \namem{cash}, \\
  &{\namem{active}(\namem{give}, \namem{Sally})}(t_4) = \top, \\
  &{\namem{using}(\namem{give}, \namem{Sally})}(t_4) = \namem{cash} \}
\end{align*}
}
The stories in the figure correspond to two trajectories, $(\ti,\tau_a)$
and $(\ti,\tau_b)$, where $\ti = [t_1,t_4]$.  States $s_1,s_2$ are
shared between the two trajectories, whereas $s_3 = \tau_a(t_4)$ and
$s_3' = \tau_b(t_4)$ differ in the values of
${\namem{using}(\namem{take},\namem{Roby})}$ and
${\namem{using}(\namem{give}, \namem{Sally})}$.  The interesting
question here is: \textit{do these trajectories represent instances of a
  trading institution?} To answer this, we construct a CSP representing
each story and verify whether or not it admits a solution.

\subsection{Adherence and the CSP}

Given the above $\inst$, $\mathcal{D}$, and $\mathcal{G}$, the sets
$A_{role}$, $B_{act}$ and $O_{art}$ are as follows:
{\small
\begin{align*}   
&A_{\namem{Buyer}} = \{\namem{Roby}\}, A_{\namem{Seller}} = \{\namem{Sally}\}, \\   
&B_{\namem{Pay}} = \{\namem{give}\}, B_{\namem{ReceiveGoods}} = \{\namem{take}\}, \\
&B_{\namem{ReceivePayment}} = \{\namem{take}\}, B_{\namem{GiveGoods}} = \{\namem{give}\}, \\
&O_{\namem{PayForm}} = \{\namem{cash}\}, O_{\namem{Goods}} = \{\namem{battery}\}.
\end{align*}

The variables of the constraint network $(\overline{W},\mathcal{C})$
obtained from $\inst$, $\mathcal{D}$, $\mathcal{G}$ and the trajectory
$(\ti,\tau_a)$ are shown in Table~\ref{table:cvals}.

\begin{table}[!ht]
  \centering
  \caption{Variables $w_{\rho,\sti}$ in $\overline W$ and their values.}
  \label{table:cvals}
  \begin{tabular}{||l|l|l||}
    \hline
    $\rho$ & $\sti$ & $w_{\rho,\sti}$ \\ \hline
    \hline
    $\namem{active}(\namem{give},\namem{Roby})$ & $[1,1]$ & $\bot$ \\ \hline
    $\namem{active}(\namem{give},\namem{Roby})$ & $[2,3]$ & $\top$ \\ \hline
    $\namem{active}(\namem{give},\namem{Roby})$ & $[4,4]$ & $\bot$ \\ \hline
    $\namem{using}(\namem{give},\namem{Roby})$ & $[2,3]$ & $\namem{cash}$ \\ \hline
    $\namem{active}(\namem{take},\namem{Roby})$ & $[1,1]$ & $\bot$ \\ \hline
    $\namem{active}(\namem{take},\namem{Roby})$ & $[4,4]$ & $\top$ \\ \hline
    $\namem{using}(\namem{take},\namem{Roby})$ & $[4,4]$ & $\namem{battery}$ \\ \hline
    $\namem{active}(\namem{give},\namem{Sally})$ & $[1,3]$ & $\bot$ \\ \hline
    $\namem{active}(\namem{give},\namem{Sally})$ & $[4,4]$ & $\top$ \\ \hline
    $\namem{using}(\namem{give},\namem{Sally})$ & $[4,4]$ & $\namem{battery}$ \\ \hline
    $\namem{active}(\namem{take},\namem{Sally})$ & $[1,2]$ & $\bot$ \\ \hline
    $\namem{active}(\namem{take},\namem{Sally})$ & $[3,3]$ & $\top$ \\ \hline
    $\namem{active}(\namem{take},\namem{Sally})$ & $[4,4]$ & $\bot$ \\ \hline
    $\namem{using}(\namem{take},\namem{Sally})$ & $[3,3]$ & $\namem{cash}$ \\ \hline
  \end{tabular}
\end{table}

The constraints reflecting the norms $N1-N9$ in our example are
constructed as follows.  For the first obligation norm $N1$, and
recalling equation (\ref{const.must}) above, we have:
\begin{align*}
  C_{N1}
  & \equiv
  \bigwedge_{a \in A_{Buyer}}
  \bigvee_{b \in B_{Pay}} 
  \exists \sti : w_{\namem{active}(b,a),\sti} = \top
  \\
  & \equiv
  \exists \sti : (w_{\namem{active}(give,Roby),\sti} = \top)
\end{align*}
Intuitively, there must be a segment in the execution trajectory during
which Roby performs the `give' behavior.
Similarly, other obligation norms induce the following
constraints:
\begin{align*}
  C_{N2}
  & \equiv
  \exists \sti : (w_{\namem{active}(take,Roby),\sti} = \top),
  \\
  C_{N3}
  & \equiv
  \exists \sti : (w_{\namem{active}(take,Sally),\sti} = \top),
  \\
  C_{N4}
  & \equiv
  \exists \sti : (w_{\namem{active}(give,Sally),\sti} = \top).
\end{align*}
%
%
%
As can be seen from Table~\ref{table:cvals}, all obligation norms are
satisfied by trajectory $(\ti,\tau_a)$ in our example.  For instance,
constraint $C_{N1}$ relative to norm $N1$ is satisfied thanks to the
fact that $w_{\namem{active}(\namem{give},\namem{Roby}),[2,3]} = \top$
(second line in Table~\ref{table:cvals}).

The constraint $C_{N5}$, induced by the usage norm $N5$, is constructed 
(recall equation~\ref{const.use}), as follows:
\begin{align*}
  C_{N5} & \equiv
  \bigwedge_{a \in A_{Buyer}}
  \bigwedge_{b \in B_{Pay}} 
  \bigvee_{o \in O_{PayForm}}
  \\
  & \hspace{3ex}
  \forall \sti : w_{\namem{active}(b,a),\sti} 
   \rightarrow 
  \left( \exists \sti' : \sti \sqsubseteq \sti'
  \wedge
  (w_{\namem{using}(b,a),\sti'} = o) \right)
  \\
  & \equiv \forall \sti : w_{\namem{active}(give,Roby),\sti} 
  \\
  & \hspace{3ex} \rightarrow 
  \left( \exists \sti' : \sti \sqsubseteq \sti'
  \wedge
  (w_{\namem{using}(give,Roby),\sti'} = cash) \right).
\end{align*}
Intuitively, the execution trajectory must be such that whenever Roby
performs the `give' behavior, it does so using `cash'.
The constraints for the other usage norms are constructed similarly:
\begin{align*}
  C_{N6} & \equiv
  \forall \sti : w_{\namem{active}(take,Roby),\sti} 
  \\
  & \hspace{3ex} \rightarrow 
  \left( \exists \sti' : \sti \sqsubseteq \sti'
  \wedge
  (w_{\namem{using}(take,Roby),\sti'} = battery) \right),
  \\
  C_{N7} & \equiv
  \forall \sti : w_{\namem{active}(take,Sally),\sti} 
  \\
  & \hspace{3ex} \rightarrow 
  \left( \exists \sti' : \sti \sqsubseteq \sti'
  \wedge
  (w_{\namem{using}(take,Sally),\sti'} = cash) \right),
  \\
  C_{N8} & \equiv
  \forall \sti : w_{\namem{active}(give,Sally),\sti} 
  \\
  & \hspace{3ex} \rightarrow 
  \left( \exists \sti' : \sti \sqsubseteq \sti'
  \wedge
  (w_{\namem{using}(give,Sally),\sti'} = battery) \right).
\end{align*}
%
%
All these constraints are also satisfied by trajectory $(\ti,\tau_a)$ in
our example.  For instance, $C_{N5}$ is satisfied because $\namem{Roby}$
activates the $\namem{give}$ behavior only in the interval $[2,3]$, and
the used object in that interval is $\namem{cash}$.  From lines 2 and 4
in Table~\ref{table:cvals}:
$w_{\namem{active}(\namem{give},\namem{Roby}),[2,3]} = \top$ and
$w_{\namem{using}(\namem{give},\namem{Roby}),[2,3]} = \namem{cash}$.

Finally, paying and receiving actions are subject to the temporal norm
$N9$, for which we construct the following constraint:
\begin{align*}
  C_{N9} & \equiv
  \bigwedge_{a \in A_{\rm Buyer}}
  \bigwedge_{b_1 \in B_{\rm Pay}} 
  \bigwedge_{b_2 \in B_{\rm ReceiveGoods}} 
  \\
  & \hspace{3ex}
  \forall \sti_1, \sti_2 :
  \left( w_{\namem{active}(b_1,a),\sti_1} = \top
  \wedge w_{\namem{active}(b_2,a),\sti_2} = \top \right)
  \\
  & \hspace{3ex}
  \rightarrow 
  \sti_1 < \sti_2
  \\
  & \equiv
  \forall \sti_1, \sti_2 :
  \left( w_{\namem{active}(give,Roby),\sti_1} = \top
  \wedge w_{\namem{active}(take,Roby),\sti_2} = \top \right)
  \\
  & \hspace{3ex}
  \rightarrow 
  \sti_1 < \sti_2
\end{align*}
where $\sti_1 = [t_1', t_1''], \sti_2 = [t_2', t_2'']$ and $\sti_1 <
\sti_2$ means $t_1'' < t_2'$.
This constraint is satisfied in our example, since
$\namem{active}(\namem{give},\namem{Roby})$ is $\top$ in $[2,3]$, and
$\namem{active}(\namem{take},\namem{Roby})$ is $\top$ in $[4,4]$ (see
Table~\ref{table:cvals}).

%




Since all constraints $C_{N1}-C_{N9}$ are satisfied, we conclude that
trajectory $(\ti,\tau_a)$ adheres to our trading institution under the
given grounding.  On the other hand, trajectory $(\ti,\tau_b)$ is
\emph{not} adherent, since it includes constraints
$w_{\namem{active}(\namem{give},\namem{Sally}),[4,4]} = \top$ and
$w_{\namem{using}(\namem{give},\namem{Sally}),[4,4]} = \namem{cash}$,
which violate constraint $C_{N8}$.  In other words, the story on the
right of Figure~\ref{fig:trade} does not represent a sound execution of
a trading institution.



%% file: robotExample.tex
\section{An Experiment with Robots and Humans}
\label{sec.rexample}

In this section, we report a proof of concept experiment that illustrates
the ability of our framework to verify and regulate norm adherence in
mixed human-robot interactions.  The experiment includes three scenarios
that stress different aspects of our framework.  The first scenario (A)
focuses on the verification problem, and features a robot observing two
humans; the second (B) and third (C) scenarios focus on the planning
problem, and respectively involve a robot and a human (B) and two robots
(C).  





\subsection{Experimental Setup}

The scenarios are realized in a smart apartment named the PEIS
Home~\citep{saffiotti2008peis} using two pepper
robots~\citep{PandeyGelin.ram2018}.  Robot's behaviors have been
implemented via NAOqi, and the environment has been customized to
simplify perception and manipulation: in particular, the objects
exchanged are sponges of different colors rather than banknotes or
batteries.  For the planning parts, we have used an off-the-shelf
timeline-based planning solution~\citep{TooCoolForSchool} which allows
us to express norm semantics in the planner's domain definition
language.

For the experiments, we have developed a simple
Institution Manager (IM) which encapsulates and implements all concepts
discussed so far.  The IM is responsible for several tasks: (1) It
realizes our institution model, allowing users to define their own
specification of institutions, domains, and groundings. (2) It checks
the admissibility of grounding by implementing the
algorithm~\ref{alg:executable}. (3) It automatically translates an
institution specification into requirements and operators in the
planning domain, to be used for verification and planning; the semantics
of norms are encoded directly into the domain definition, thus allowing
the planner to directly generate adherent plans (see
Section~\ref{subsec.planning}). (4) It controls plan execution by
dispatching behaviors to the robots.  (5) It uses feedback from the
robots to verify the adherence of plan execution to norms.  The code of
the IM, together with the complete domain specification used by the
planner are available at \url{http://aass.oru.se/~sntc/TradeExperiment} 
and videos of the scenario executions at \url{https://youtu.be/OIMy4UyJFws}.

\subsection{Institution and Domain}

The used institution specification is the same as the one in
section~\ref{sec.reasoning_example}.  The domain, however, is different.
The following domain models the relevant parts of the physical setup
used in our tests:
\begin{align*}
  A = \{&\namem{pepper_{1}}, 
         \namem{pepper_{2}}, 
         \namem{human_{1}}, 
         \namem{human_{2}},
         \dots \} \\
  B = \{&\namem{pick}, \namem{give}, \dots \} \\
  O = \{&\namem{sponge_{yellow}}, 
         \namem{sponge_{blue}}, 
         \namem{sponge_{red}},
         \dots \} \\
  F = \{&\stm{pepper_{x}}{give}{sponge_{color}}, \\
        &\stm{pepper_{x}}{pick}{sponge_{color}},
         \dots \} \\
  R = \{&\namem{active}(\namem{pepper_{x}},\namem{give}), 
         \namem{active}(\namem{pepper_{x}}, \namem{pick}),\\
        &\namem{active}(\namem{human_{x}},\namem{give}), 
         \namem{active}(\namem{human_{x}}, \namem{pick}),
         \dots \}.
\end{align*}
%
We use $x$ and $\text{color}$ subscripts as a compact way to
indicate that some affordances or state-variables apply to several
agents or objects.  As specified by the affordance relation $F$, any
robot can \name{pick} and \name{give} any \name{sponge_{color}}.  We
have omitted from $F$ the affordances of humans, who are assumed to be
capable of all behaviors.

The value of the \name{active(a,b)} state variable is the object used by
agent $a$ to run behavior $b$, or $\bot$ if $a$ is not running $b$.
This variable conglomerates the two separate variables
\name{active(a,b)} and \name{use(a,b)} that we used in
\ref{sec.reasoning_example} above.  Thus, we can now simultaneously
check the satisfaction of the `must' and the `use' norms.

%

\begin{figure}
\centering
\includegraphics[width=\linewidth]{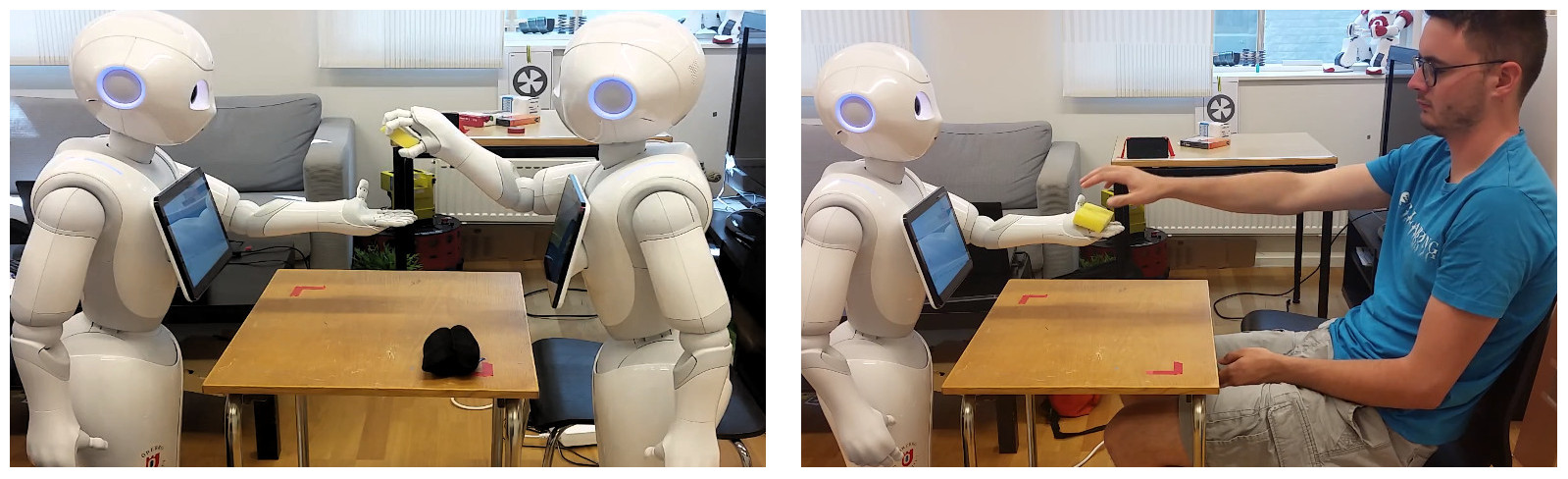}
\caption{Two agents exchange an object in a trade institution.  Scenario
  A (left): the seller and buyer roles are both grounded in robots;
  Scenario B (right): the seller is grounded in a human.}
\label{fig:twoPepper}
\end{figure}

\subsection{Scenario A}

The task is to check whether an observed execution adheres to an
institution or not.  We placed one Pepper robot in front of a table
where two humans interact, and we developed a simple ad-hoc behavior
recognizer for it which returns values of the state variables from the
video stream.  The grounding $\grnd$ is provided \textit{a-priori}:
\begin{align*}
  \grnd_A = \{& (\namem{Buyer},\namem{human_{1}}), 
                (\namem{Seller},\namem{human_{2}}) \} \\
  \grnd_B = \{&(\namem{ReceiveGoods},\namem{pick}), 
               (\namem{ReceivePayment},\namem{pick}), \\
              &(\namem{Pay},\namem{give}), (\namem{GiveGoods},\namem{give}) \} \\
  \grnd_O = \{&(\namem{PayForm},\namem{sponge_{yellow}}), 
               (\namem{Goods},\namem{sponge_{blue}}) \}
\end{align*}

\begin{figure}[!ht]
  \centering
  \includegraphics[width=\linewidth]{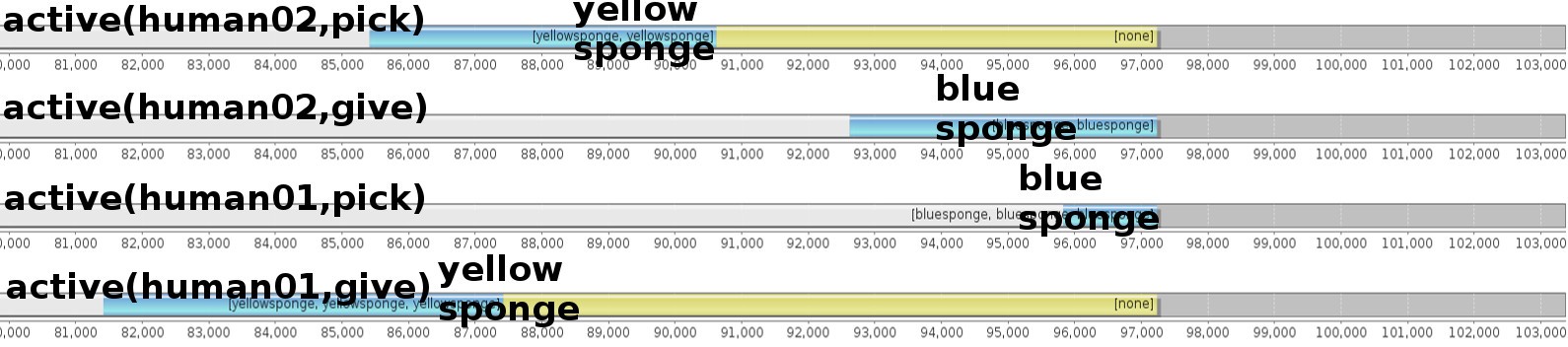}
  \caption{An observed trajectory $(\ti,\tau_1)$ from the experiment.
    This trajectory adheres to a trading institution. }
  \label{fig:AdherentT}
\end{figure}

Figure~\ref{fig:AdherentT} shows an observed trajectory $(\ti,\tau_1)$,
represented as evolution in time of values of state variables.  Here,
$\namem{human_{1}}$ gives a yellow sponge (PayForm) before it picks a
blue sponge (Goods).  All norms are satisfied, and therefore
$(\ti,\tau_1)$ is a sound (adherent) execution of the trade institution.



\begin{figure}[!ht]
  \centering
  \includegraphics[width=\linewidth,frame]{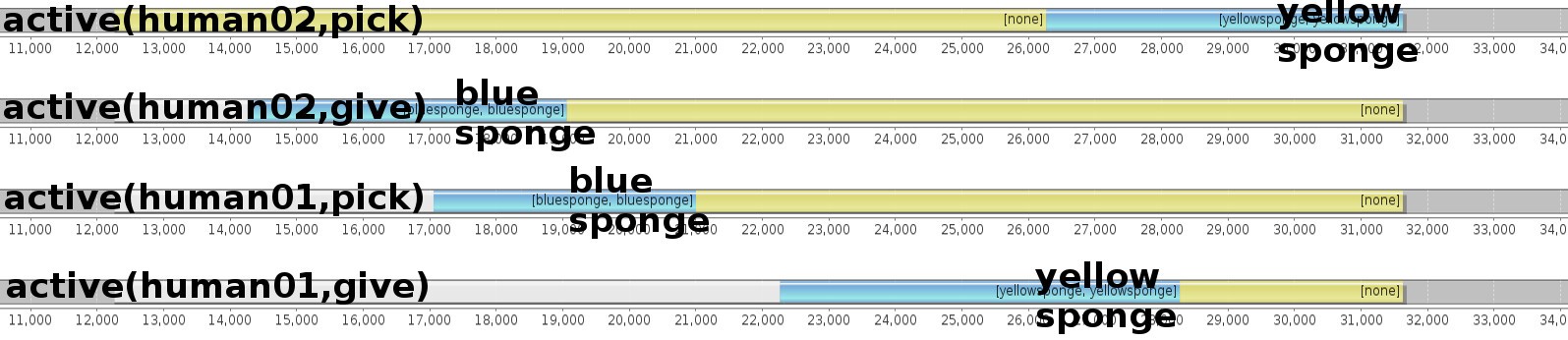}
  \caption{Another observed trajectory $(\ti,\tau_2)$.  This trajectory
    does not adhere to a trading institution. }
  \label{fig:NotAdherentT}
\end{figure}

By contrast, in the observed trajectory $(\ti,\tau_2)$, shown in
Figure~\ref{fig:NotAdherentT}, $\namem{human_{1}}$ picks a blue sponge
(Goods) before giving a yellow sponge (PayForm), instead of first paying
and then receiving the Goods. Thus, the temporal norm 'before' is not
satisfied over this trajectory, and $(\ti,\tau_1)$ is not a sound
execution of the trade institution.

\subsection{Scenario B}

We now switch our attention from verification to planning and show
that we can generate a trajectory for a robot to interact with a human
while following the norms of a trading institution. The advantage of
having modeled the scenario as an institution becomes evident when we
want to change actors and/or use different merchandise.  We replaced
one of the two humans in the previous scenario by a Pepper robot and
changed merchandise with a black battery box by changing the
grounding $\grnd_A$ and $\grnd_O$ by:
\begin{align*}
  \grnd_A' = \{& (\namem{Seller},\namem{human_{1}}), (\namem{Buyer},\namem{pepper_{1}}) \} \\
  \grnd_O' = \{& (\namem{PayForm},\namem{sponge_{yellow}}), (\namem{Goods},\namem{battery}) \}
\end{align*}
The institution specification is translated by IM, into the set of operators
and requirements for the planning domain used by the robots.  An
adherent trajectory is synthesized by the timeline-based planner whose
domain contains the semantic models of all norms. The resulting plan was
dispatched to \name{pepper_{1}}.

\subsection{Scenario C}

In the final scenario, we used two robots instead of a robot and a
human. This is achieved by simply replacing the roles grounding in
$\grnd$ by the following:
\begin{align*}
  \grnd_A'' = \{&(\namem{Buyer},\namem{pepper_{1}}), 
                 (\namem{Seller},\namem{pepper_{2}}) \}
\end{align*}
In this case, the resulting plan was dispatched to both robots:
\name{pepper_{1}} and \name{pepper_{2}}.

The grounding in all scenarios is admissible because it grounds the
roles of traders to robots that are capable of executing the
appropriate \name{give} and \name{pick} behaviors.





%% file: conclusion.tex
\section{Conclusion and Future Work}
\label{sec.conclusion}

In this work, we have introduced a theoretical framework for modeling and
reasoning about norms in robotics.  The framework is grounded on the
notion of institution, which provides a way to model how agents should
behave in a given social context.  The framework distinguishes between
abstract norms and their instantiation into a concrete domain, and it
combines insights from the fields of multi-agent systems and of
robotics.
It enables the definition of relevant computational tasks, such as
verification and planning.  Notably, the framework provides support for
artifacts, which is of high importance in robotics, since robots and
humans interact and coordinate via relevant objects in the environment.
We have shown how the verification problem can be cast to a CSP, and how
this enables the use of constraint-based planning and plan execution
technology to control a robot system.
%
Finally, we have demonstrated our framework in a physical system
comprising both humans and robots.  Additional experiments using
different institutions and mixed human-robot interactions can be found
in~\cite{tomic2018towards}.

Institutions encapsulate all norms, roles, and artifacts that are relevant in a given social context. However, it is often the case in human societies that several institutions are relevant simultaneously, --- e.g., children who participate in a game playing institution still adhere to the norms of the ‘surrounding’ school institution. Relating different institutions and exploit these relations in verification and planning is the topic of ongoing work. Also, we plan to study how to use institutions for goal reasoning --- e.g., if a robot realizes it needs a new battery, it may decide that the best way to obtain it is
to engage in a trading institution.
